\documentclass[10pt,twocolumn,letterpaper]{article}

\usepackage[pagenumbers]{cvpr} 

\usepackage[dvipsnames]{xcolor}

\definecolor{cvprblue}{rgb}{0.21,0.49,0.74}
\usepackage[pagebackref,breaklinks,colorlinks,citecolor=cvprblue]{hyperref}

\def\paperID{xxx} 
\def\confName{CVPR}
\def\confYear{2024}
\definecolor{forestgreen}{RGB}{47, 159, 87}
\newcommand{\cmark}{\color{forestgreen}\ding{51}}%
\newcommand{\xmark}{\color{red}\ding{55}}%
\usepackage{pifont}
\usepackage{makecell}
\newlength\savewidth\newcommand\shline{\noalign{\global\savewidth\arrayrulewidth
  \global\arrayrulewidth 1pt}\hline\noalign{\global\arrayrulewidth\savewidth}}
\newcommand{\tablestyle}[2]{\setlength{\tabcolsep}{#1}\renewcommand{\arraystretch}{#2}\centering\small}
\def\eg{\emph{e.g.}}
\def\ie{\emph{i.e.}}
\definecolor{lightgrey}{RGB}{244,244,244}

\title{Hierarchical Spatio-temporal Decoupling for Text-to-Video Generation }

\author{Zhiwu Qing$^{1}$ \quad Shiwei Zhang$^{2*}$ \quad Jiayu Wang$^{2}$ \quad Xiang Wang$^1$  \\ Yujie Wei$^{3}$ \quad Yingya Zhang$^2$    \quad Changxin Gao$^{1*}$ \quad Nong Sang$^{1}$ \\
$^1$Key Laboratory of Image Processing and Intelligent Control \\ School of Artificial Intelligence and Automation, Huazhong University of Science and Technology \\ 
$^2$Alibaba Group \quad $^3$Fudan University \\ 
{\tt\small \{qzw, wxiang, cgao, nsang\}@hust.edu.cn} \\
{\tt\small\{zhangjin.zsw, wangjiayu.wjy, yingya.zyy\}@alibaba-inc.com } \\
{\tt\small yjwei22@m.fudan.edu.cn}
\vspace{-0.6cm}
}

\newcommand{\method}{\texttt{HiGen}\xspace}

\usepackage{colortbl} 
\usepackage{xcolor}

\begin{document}
\maketitle
\let\thefootnote\relax\footnotetext{$^*$Corresponding authors.}
\let\thefootnote\relax\footnotetext{Project page: \url{https://higen-t2v.github.io/}.}
\begin{abstract}

Despite diffusion models having shown powerful abilities to generate photorealistic images, generating videos that are realistic and diverse still remains in its infancy. 
One of the key reasons is that current methods intertwine spatial content and temporal dynamics together, leading to a notably increased complexity of text-to-video generation (T2V).
In this work, we propose \method, a diffusion model-based method that improves performance by decoupling the spatial and temporal factors of videos from two perspectives, i.e., structure level and content level.
At the structure level, we decompose the T2V task into two steps, including spatial reasoning and temporal reasoning, using a unified denoiser. 
Specifically, we generate spatially coherent priors using text during spatial reasoning and then generate temporally coherent motions from these priors during temporal reasoning.
At the content level, we extract two subtle cues from the content of the input video that can express motion and appearance changes, respectively.
These two cues then guide the model's training for generating videos, enabling flexible content variations and enhancing temporal stability.
Through the decoupled paradigm, \method can effectively reduce the complexity of this task and generate realistic videos with semantics accuracy and motion stability.
Extensive experiments demonstrate the superior performance of \method over the state-of-the-art T2V methods.

\end{abstract}    
\vspace{-0.3cm}
\section{Introduction}
\label{sec:intro}

The purpose of text-to-video generation (T2V) is to generate corresponding photorealistic videos based on input text prompts.
These generated videos possess tremendous potential in revolutionizing video content creation, particularly in movies, games, entertainment short videos, and beyond, where their application possibilities are vast.
Existing methods primarily tackle the T2V task by leveraging powerful diffusion models, leading to substantial advancements in this domain.

\begin{figure}[t]
    \centering
    \includegraphics[width=1.0\linewidth]{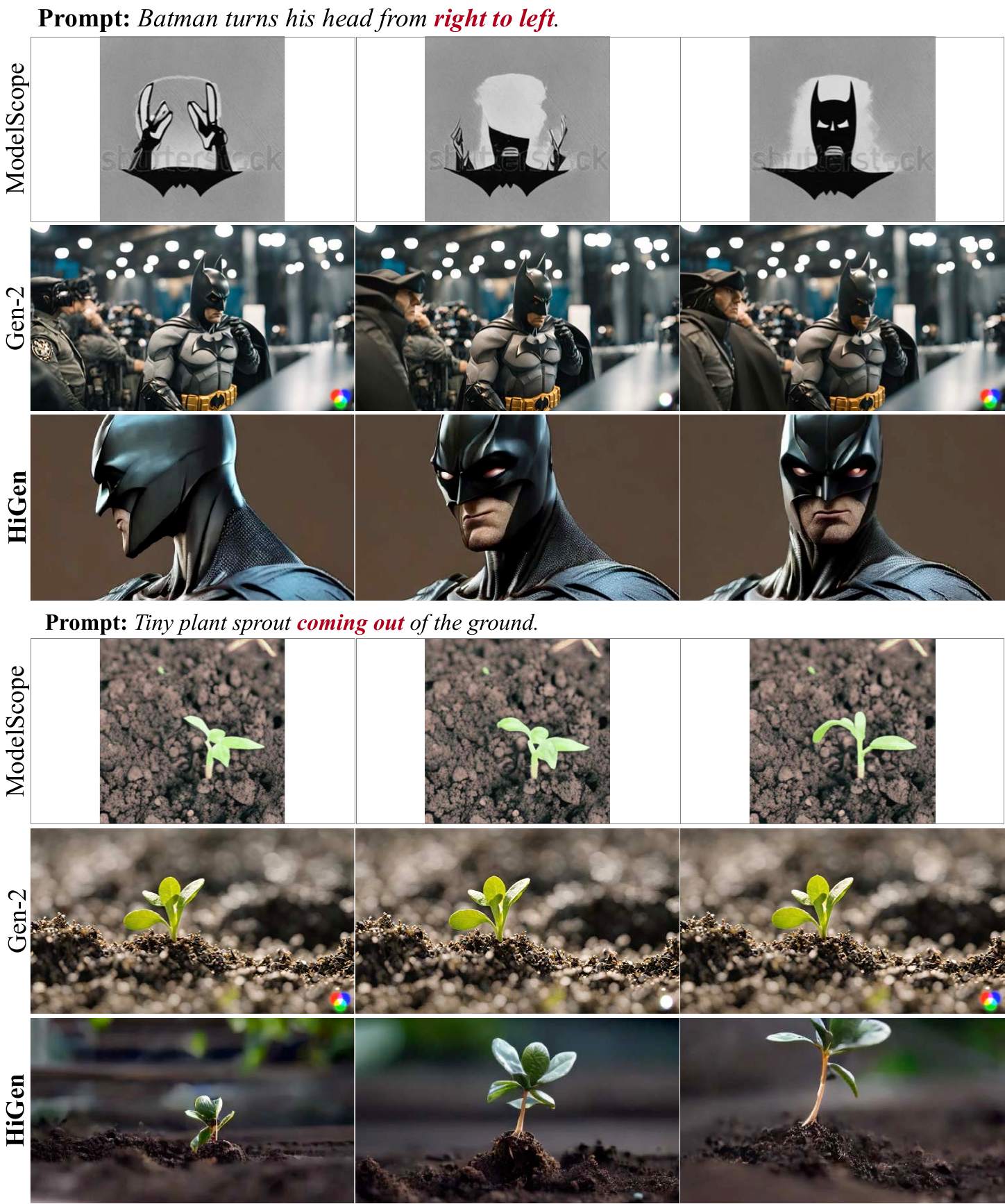}
    \vspace{-7mm}
    \caption{
    Visual comparison with ModelScopeT2V~\cite{wang2023modelscope} and Gen-2~\cite{esser2023gen1}. 
    The videos generated by ModelScopeT2V exhibit noticeable motion but suffer from lower spatial quality. 
    However, while Gen-2 produces realistic frames, they are mostly static with minimal motion. 
    In contrast, the results of our \method demonstrate both realistic frames and rich temporal variations.
    }
    \vspace{-6mm}
    \label{fig:first_fig}
\end{figure}

\begin{figure*}[t]
    \centering
    \includegraphics[width=0.99\linewidth]{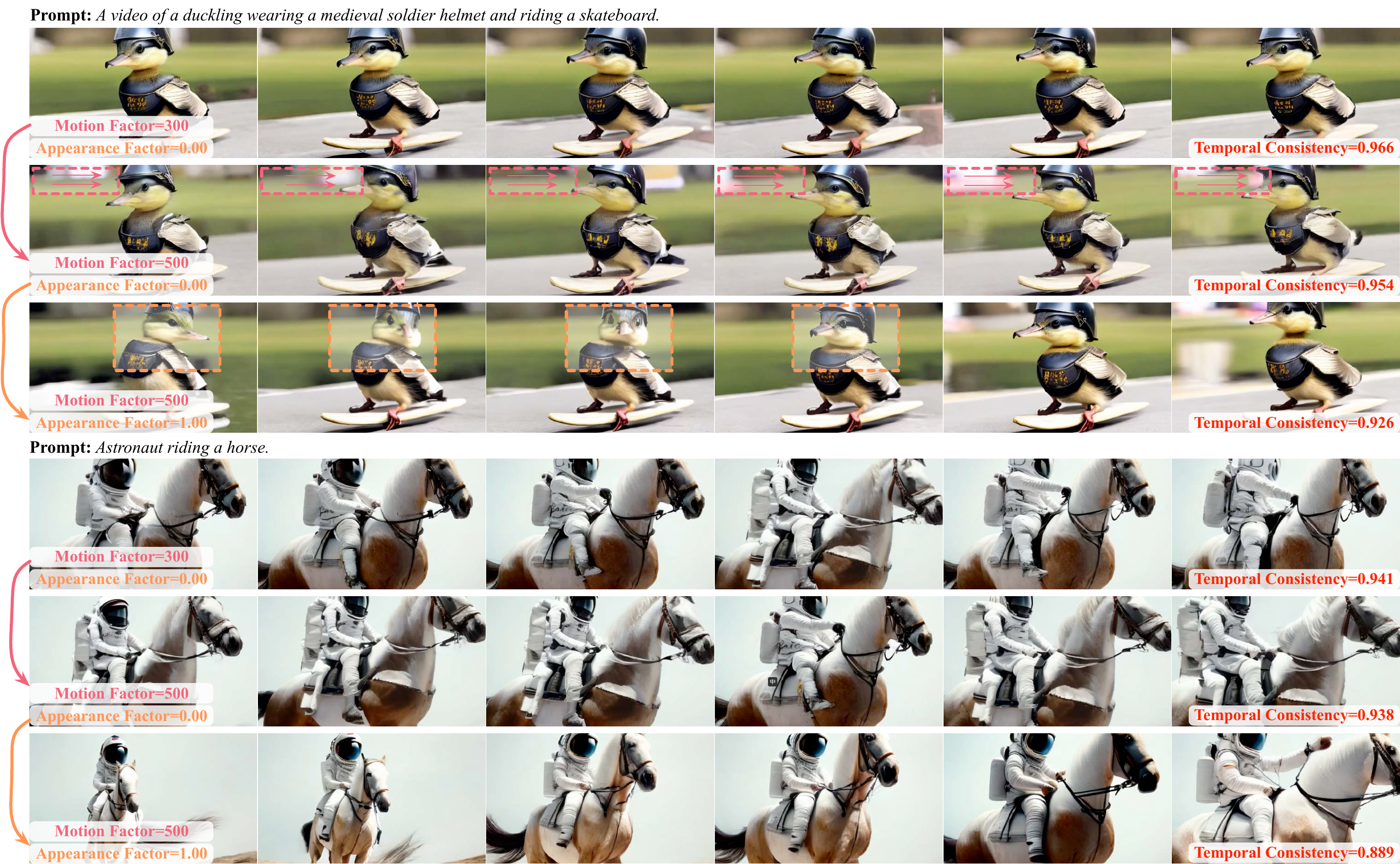}
    \vspace{-3mm}
    \caption{
    {The impact of motion factors and appearance factors.} 
    Larger motion factors introduce dynamic motions to the videos instead of static scenes, while larger appearance factors contribute to richer temporal semantic variations in the generated videos.
    }
    \vspace{-6mm}
    \label{fig:sec_fig}
\end{figure*}

Typically, mainstream approaches~\cite{blattmann2023videoLDM,ho2022imagen_video,wang2023videofactory,wang2023modelscope,chen2023videocrafter1} attempt to generate videos by extending text-to-image (T2I) models by designing suitable 3D-UNet architectures.
However, due to the complex distribution of high-dimensional video data, directly generating videos with both realistic spatial contents and diverse temporal dynamics jointly is in fact exceedingly challenging, which often leads to unsatisfactory results produced by the model.
For example, as shown in Fig.~\ref{fig:first_fig}, videos generated by ModelScopeT2V~\cite{wang2023modelscope} exhibit dynamics but suffer from lower spatial quality. 
Conversely, videos from Gen-2~\cite{esser2023gen1}  showcase superior spatial quality but with minimal motions.
On the other hand, VideoFusion~\cite{luo2023videofusion} considers spatial redundancy and temporal correlation from the noise perspective by decomposing input noise into base noise and residual noise.
However, it remains challenging to directly denoise videos with spatio-temporal fidelity from the noise space.

\begin{figure*}[t]
    \centering
    \includegraphics[width=1.0\linewidth]{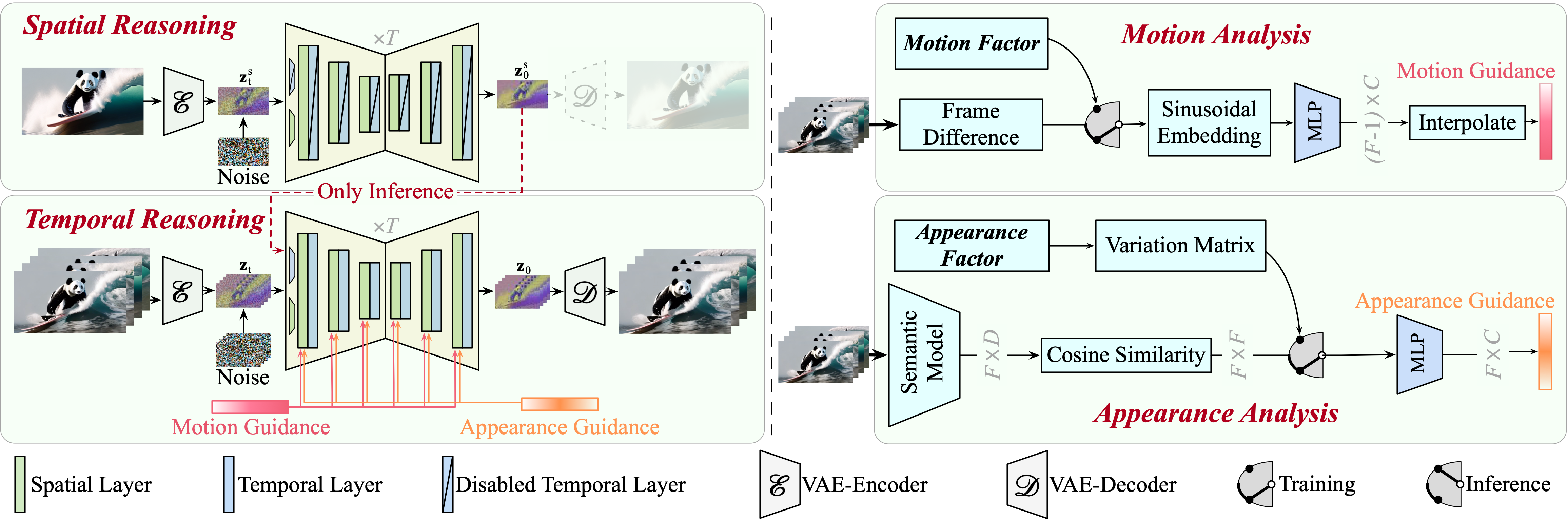}
    \vspace{-7mm}
    \caption{{The overall framework of \method.} \textit{Left:} 
    The structure-level spatio-temporal decoupling. Firstly, spatial reasoning is performed to obtain latent embeddings of spatial priors. Then, these spatial priors are used for temporal reasoning to generate videos.
    \textit{Right:} 
    The content-level motion-appearance decoupling. Motion analysis and appearance analysis refer to the calculations of motion and appearance guidance, respectively.
    }
    \vspace{-6mm}
    \label{fig:framework}
\end{figure*}

Based on the above observations, we propose a new diffusion model-based \method approach that decouples videos into spatial and temporal factors from two perspectives, namely structure level and content level.
For the structure level, in light of the separability of space and time~\cite{feichtenhofer2016_2stream,wang2016tsn} in video data, we decompose the T2V task into distinct spatial reasoning and temporal reasoning processes, all predicated on a unified model.
During spatial reasoning, we utilize text prompts to generate spatial priors that are semantically coherent. 
These priors are then used in temporal reasoning to generate temporally coherent motions.
For the content level, we extract two cues that respectively represent the motion and appearance variations in videos and utilize them as conditions for training the model.
By this means, we can enhance the stability and diversity of generated videos by flexibly controlling the spatial and temporal variations through manipulating the two conditions, as shown in Fig.~\ref{fig:sec_fig}.
Thanks to this hierarchically decoupled paradigm, \method ensures simultaneous high spatial quality and motion diversity in the generated videos.
To validate \method, we extensively conduct qualitative and quantitative analyses, comparing it with state-of-the-art methods on the public dataset, \ie, MSR-VTT~\cite{xu2016msrvtt}.
The experimental results demonstrate the effectiveness of \method and its superior performance compared to current methods.

\section{Related Works}
\label{sec:related_works}

\textbf{Diffusion-based Text-to-Image Generation.}
Recently, diffusion models have greatly advanced the progress of text-driven photorealistic image synthesis. 
Initially, due to the substantial computational burden associated with performing iterative denoising on high-resolution images, early works~\cite{ho2020diffusion,song2020diffusion} predominantly concentrated on the generation of low-resolution images.
To generate high-resolution images, a series of methods~\cite{ramesh2022unclip,saharia2022imagen,ho2022cascadedcdm,nichol2021glide,balaji2022ediff} have employed super-resolution techniques on low-resolution images, while others~\cite{rombach2022stablediffusion,gu2022vqdiffusion,podell2023sdxl} have utilized decoders to decode features from the latent space. 
Besides, exploring how to achieve flexible and controllable image generation is also an important research direction, such as ControlNet~\cite{zhang2023controlnet}, Composer~\cite{huang2023composer}, DreamBooth~\cite{ruiz2023dreambooth}, \textit{etc}.
Building upon state-of-the-art image generation methods, numerous advanced video generation~\cite{guo2023animatediff,zhang2023show1} or editing~\cite{wu2023tune_a_video,qi2023fatezero,molad2023dreamix,ceylan2023pix2video,bar2022text2live,zhao2023motiondirector} approaches have been developed by fine-tuning with additional temporal transformer layers or controlling the inference process. 
In this work, we fine-tune a high-quality text-to-video model by leveraging the powerful and efficient text-to-image model, \textit{i.e.,} Stable Diffusion~\cite{rombach2022stablediffusion}.

\noindent
\textbf{Diffusion-based Text-to-Video Generation.} Video synthesis methods strive to explore the generation of temporally coherent videos. 
Early works primarily relied on Generative Adversarial Networks (GANs)~\cite{yu2022video_implicit_GAN,skorokhodov2022stylegan-v,hong2022cogvideo,zhao2018learning,gan1,mostgan,gan3,tian2021mocoganhd}. 
Recently, breakthroughs have been achieved through diffusion-based methods, which can be broadly categorized into two paradigms: 
\textit{(i)} introducing additional temporal layers~\cite{blattmann2023videoLDM,he2022videolvdm,wu2023tune_a_video,wang2023videofactory,guo2023animatediff,zhou2022magicvideo,wang2023modelscope,luo2023videofusion,ge2023pyoco,liu2023dsdn,xing2023simda} or operations~\cite{an2023latent_shift} for fine-tuning. 
%
To reduce the complexity of video generation, some works~\cite{singer2022make_a_video,ho2022imagen_video,wang2023lavie,zhang2023show1,blattmann2023videoLDM,zhou2022magicvideo,li2023videogen} employ a series of big diffusion models for generating and upsampling videos given the input text. 
Besides, another line~\cite{luo2023videofusion,ge2023pyoco} alleviates the training difficulty by increasing temporal correlations between frame-wise noise, but this may limit the temporal diversity of the generated videos. 
\textit{(ii)} Controlling the inference process through training-free designs~\cite{khachatryan2023text2video_zero,hong2023direct2v,huang2023freebloom,duan2023diffsynth,lian2023lvd}. 
This paradigm does not require training but typically yields lower temporal continuity compared to fine-tuning-based methods. 

Unlike existing approaches, in this work, we explore a hierarchical spatio-temporal decoupling paradigm based on the more promising fine-tuning strategy to train T2V models that exhibits both rich temporal variations and high-quality spatial content.

\section{Approach}

\subsection{Preliminaries}

\noindent
In this work, we use $\mathbf{x}_0=[\mathbf{x}_0^1, \dots, \mathbf{x}_0^F]$ to denote a video with $F$ frames.
Following Stable Diffusion~\cite{rombach2022stablediffusion}, we map the video frames into the latent space by a Variational Auto-Encoder (VAE)~\cite{kingma2013vae} as $\mathbf{z}_0=[\mathcal{E}(\mathbf{x}_0^1), \dots, \mathcal{E}(\mathbf{x}_0^F)]$, where $\mathcal{E}$ denotes the encoder, and $\mathbf{z}_0$ can be decoded by the decoder $\mathcal{D}$ to reconstruct RGB pixels. 
With the video latent embedding $\mathbf{z}_0$, the diffusion process involves gradually add random noises into $\mathbf{z}_0$ using a $T$-Step Markov chain~\cite{kong2021fast}:
\begin{equation}
    q(\mathbf{z}_{t}|\mathbf{z}_{t-1}) = \mathcal{N}(\mathbf{z}_{t};\sqrt{1-\beta_{t-1}}\mathbf{z}_{t-1}, \beta_{t}I),
\end{equation}
where $\beta_{t}$ refers to the noise schedule, and $\mathcal{N}(\cdot; \cdot)$ indicates the Gaussian noise.
After being corrupted by noise, the obtained $\mathbf{z}_t$ is fed into a 3D-UNet for noise estimation, enabling progressive denoising process to restore a clean video latent embedding.

In both the training and inference phase of the 3D-UNet,
we adopt the same approach as in Stable Diffusion to inject the text condition and diffusion time $t$ separately into the spatial Transformer layer and residual block. 
For brevity, we omit the details of these two components in Fig.~\ref{fig:framework}.

\subsection{Structure-level Decoupling}
\label{sec:decoupled}

\noindent
From a model structure perspective, we divide the T2V generation into two steps: spatial reasoning and temporal reasoning. 
Spatial reasoning aims to maximize the utilization of the knowledge in T2I models, thereby providing high-quality spatial priors for temporal reasoning. 
To obtain this prior, we employ the same textual-conditional image synthesis procedure like Stable Diffusion~\cite{rombach2022stablediffusion}. 
Specifically, as shown in the \textit{Spatial Reasoning} card in Fig.~\ref{fig:framework}, we only leverage the spatial layers in 3D-UNet while disregarding its temporal components for spatial generation. 
After $T$ steps of denoising, the spatial prior is represented as ${\mathbf{z}}_0^\text{s}$.
It is worth noting that ${\mathbf{z}}_0^\text{s}$ does not need to be decoded by the VAE decoder $\mathcal{D}$ to reconstruct its pixel values. 
This allows for an efficient input of ${\mathbf{z}}_0^\text{s}$ into the subsequent temporal reasoning.

\begin{figure}[t]
    \centering
    \includegraphics[width=0.95\linewidth]{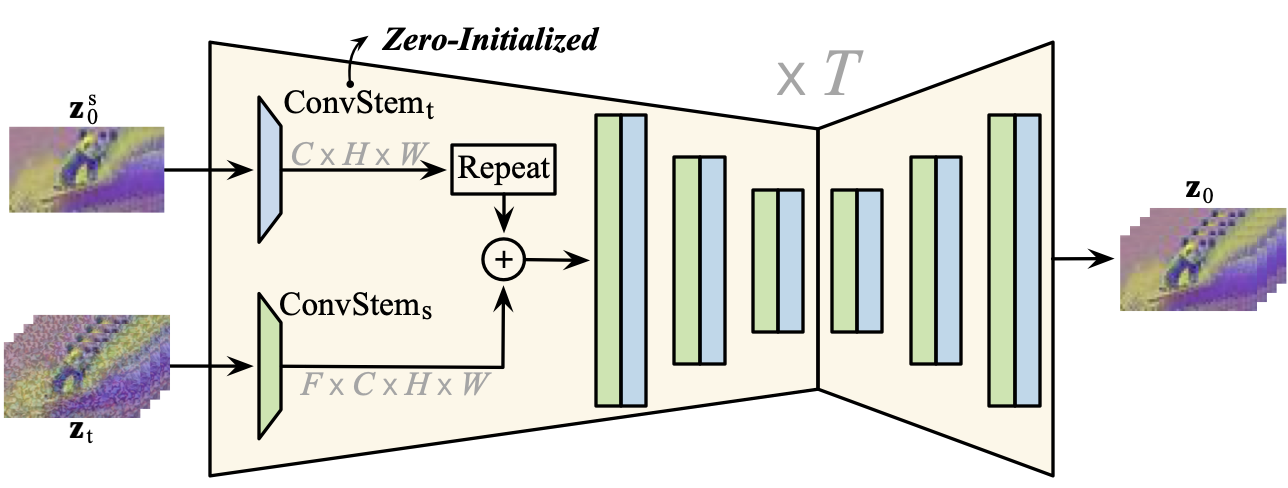}
    \vspace{-3mm}
    \caption{{The spatial prior for temporal reasoning.}
    }
    \vspace{-6mm}
    \label{fig:merge_st}
\end{figure}

The core idea of temporal reasoning is to synthesis diverse temporal dynamics for video generation on top of the spatial prior ${\mathbf{z}}_0^\text{s}$.
Specifically, as shown in shown in Fig.~\ref{fig:merge_st}, we initialize a convolutional layer with all zeros (\ie, $\text{ConvStem}_\text{t}(\cdot)$) for  ${\mathbf{z}}_0^\text{s}$ separately.
The structure of $\text{ConvStem}_\text{t}(\cdot)$ is identical to the image pre-trained convolutional stem in the UNet (\ie, $\text{ConvStem}_\text{s}(\cdot)$). 
After passing through $\text{ConvStem}_\text{t}(\cdot)$, we repeat the spatial prior $F$ times and add it to the noisy video embeddings ${\mathbf{z}}_t$ for UNet.

Besides, we further clarify some details of the proposed structure-level decoupling from the following three aspects:
\textit{(i)} Merging the spatial prior after the first convolutional stem enables effective guidance for all the spatial and temporal layers in the 3D-UNet, which maximizes the utilization of the rich semantic priors present in the spatial prior.
\textit{(ii)} Our temporal reasoning and spatial reasoning share the same spatial layers. This allows the temporal reasoning phase to leverage the pre-trained knowledge in the spatial layers, facilitating more accurate temporal synthesizing.
\textit{(iii)} The temporal layers consist of a series of temporal convolutions and temporal self-attention layers following~\cite{wang2023modelscope}. 
Despite similar structures, our temporal layers can be freed from intricate spatial contents and can solely focus on generating fine-grained temporal motions between consecutive frames, as demonstrated in Fig.~\ref{fig:st_decoupling}.

\subsection{Content-level Decoupling}
\label{sec:motion_semantic}

\begin{figure}[t]
    \centering
    \includegraphics[width=1.0\linewidth]{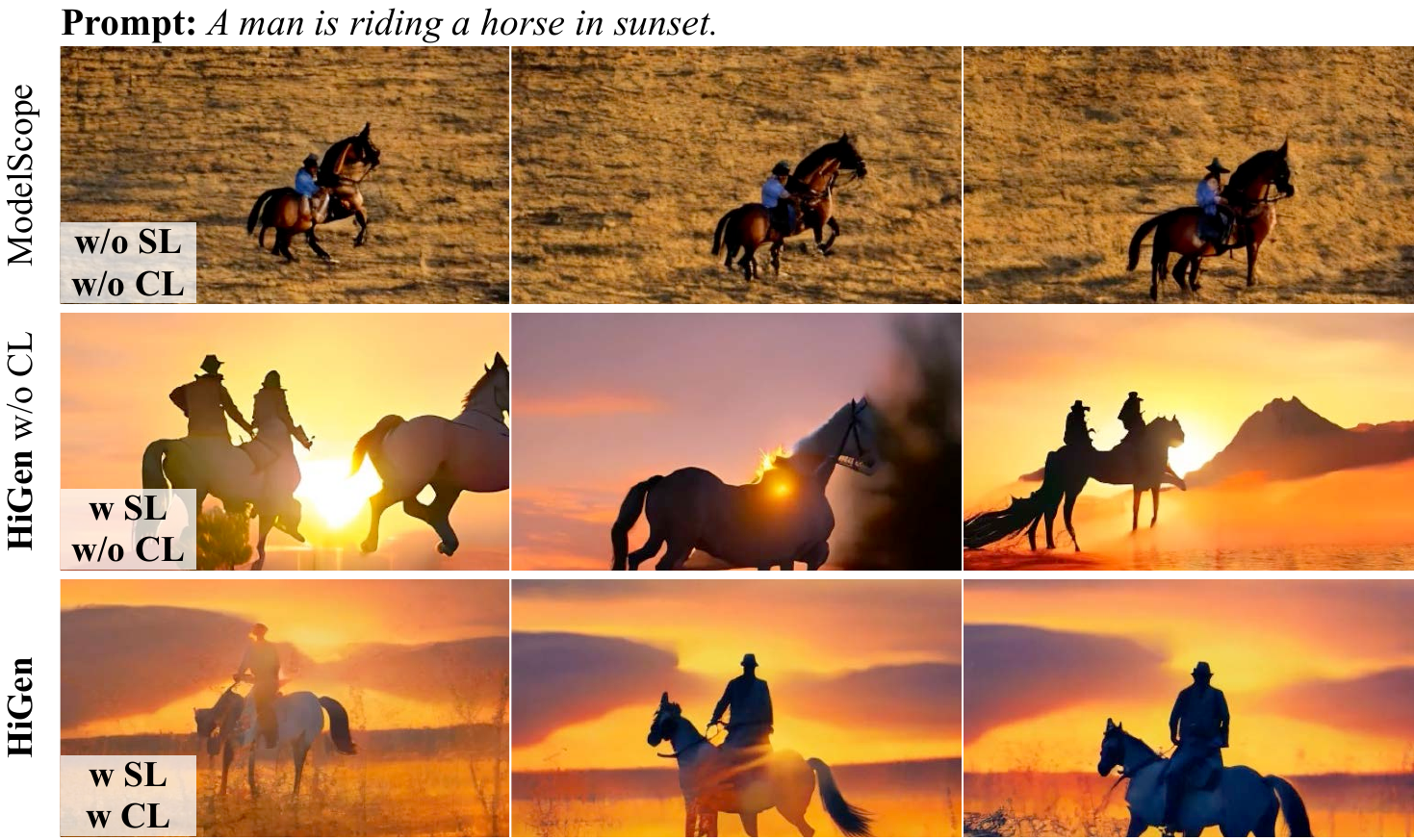}
    \vspace{-7mm}
    \caption{Visualization for the effect of Structure-Level (SL) decoupling and Content-Level (CL) decoupling.}
    \vspace{-5mm}
    \label{fig:abla1}
\end{figure}

Based on the structure-level decoupling, our paradigm is already capable of generating spatially realistic frames.
However, in the temporal case, it still faces two challenges:
nearly static video frames (\eg, Gen-2~\cite{esser2023gen1}) and unstable temporal variations (\eg, the 2nd row in Fig.~\ref{fig:abla1}).
Hence, we further propose motion and appearance decoupling for video content level to enhance the vividness and stability of synthesized videos.

\noindent
\textbf{Motion Analysis.} For motion decoupling, we present motion analysis to quantify the magnitude of motion between frames, providing motion guidance for 3D-UNet.
FPS (frames per second), which reflects the playback speed of the video, may seem like an intuitive choice~\cite{zhou2022magicvideo}. 
However, FPS alone does not accurately reflect the motion in a video (\eg, static videos may also have a high FPS). 
Inspired by video understanding tasks~\cite{wang2016tsn,zhi2021mgsampler}, frame differencing with negligible computational cost is an effective method for measuring video motion.
Therefore, for a sequence of $F$ frames, we define the \textit{\textbf{motion factor}} as $\gamma^\text{m}_{f}=||\mathbf{z}_0^f-\mathbf{z}_0^{f+1}||$, which indicates the magnitude of the pixel differences between adjacent frames.
By computing $\gamma^\text{m}_{f}$ for $F$ frames, we can obtain $F-1$ motion factors: $\tilde{\mathbf{r}}^\text{m} = [\gamma^\text{m}_{1}, \dots, \gamma^\text{m}_{F-1}]\in \mathbb{R}^{F-1}$.

To incorporate $\tilde{\mathbf{r}}^\text{m}$ into the 3D-UNet, we first round $\gamma^\text{m}_{f}$ and then utilize sinusoidal positional encoding~\cite{attention_is_all_u_need} and a zero-initialized MLP (Multi-Layer Perceptron) to map it into a $C$-dimensional space:
\begin{equation}
\label{eq:rm}
    \mathbf{r}^\text{m}=\text{Interpolate}(\text{MLP}(\text{Sin}(\text{Round}(\tilde{\mathbf{r}}^\text{m}))))\in \mathbb{R}^{F\times C},
\end{equation}
where $\text{Interpolate}(\cdot)$ is a linear interpolation function that aligns the $F-1$ motion factors with the actual number of frames (\ie, $F$). 
Next, the motion guidance $\mathbf{r}^\text{m}$ is added to the time-step embedding vector of the diffusion sampling step $t$~\cite{ho2020diffusion}. 
Therefore, $\mathbf{r}^\text{m}$ is integrated with features in each residual block.

\noindent
\textbf{Appearance Analysis.} 
The motion factor describes pixel-level variations between adjacent frames while it cannot measure the appearance changes. 
To address this, we leverage existing visual semantic models such as, DINO~\cite{caron2021dino,oquab2023dinov2}, CLIP~\cite{radford2021CLIP}, for appearance analysis between frames:
\begin{equation}
\label{eq:rs}
    \begin{aligned}
        \mathbf{g}=\text{Norm}(\Omega(\mathbf{x}_0))\in \mathbf{R}^{F\times D}, 
        \tilde{\mathbf{r}}^{\text{a}}=\mathbf{g}\otimes \mathcal{T}(\mathbf{g}) \in \mathbb{R}^{F\times F},
    \end{aligned}
\end{equation}
where $\Omega(\cdot)$ and $\text{Norm}(\cdot)$ refer to the semantic model and normalization operation, respectively. 
$\otimes$ is matrix multiplication, and $\mathcal{T}(\cdot)$ means the transpose operation. 
Therefore, $\tilde{\mathbf{r}}^{\text{a}}$ represents the cosine similarities between all frames, which is then transformed using a zero-initialized MLP to obtain the appearance guidance: $\mathbf{r}^{\text{a}}=\text{MLP}(\tilde{\mathbf{r}}^{\text{a}})\in \mathbb{R}^{F\times C}$. 
Afterwards, $\mathbf{r}^{\text{a}}$ is inputted into the 3D-UNet in the same way as the motion guidance $\mathbf{r}^\text{m}$.

In general, a video clip with large appearance variations will have a lower cosine similarity value between the first and last frames, \ie, $\tilde{\mathbf{r}}^{\text{a}}_{0, F-1}$. 
To align with intuition, we further define the \textit{\textbf{appearance factor}} as $\gamma^{\text{a}}=1 - \tilde{\mathbf{r}}^{\text{a}}_{0, F-1}$. 
In this case, a larger appearance factor $\gamma^{\text{a}}$ corresponds to significant appearance variations in the generated videos.
In training, we calculate the appearance guidance from real videos using Eq.~\ref{eq:rs}.
In inference, we manually construct the variation matrix ($\tilde{\mathbf{r}}^\text{a}$) based on the appearance factor $\gamma^{\text{a}}$, which will be discussed in the next section.

\subsection{Training and Inference}
\label{sec:training_infer}

\textbf{Training.} 
We train our 3D-UNet through image-video joint training~\cite{ho2022img_vid_joint,wang2023videocomposer}. 
Specifically, we allocate one-fourth of the GPUs for image fine-tuning (\ie, spatial reasoning), while the remaining GPUs are utilized for video fine-tuning (\ie, temporal reasoning). 
For image GPUs, we only optimize the spatial parameters that were pre-trained by Stable Diffusion~\cite{rombach2022stablediffusion} to preserve its spatial generative capability.
On the other hand, for video fine-tuning, we optimize all parameters based on strong spatial priors.
To ensure efficiency, we utilize the middle frame of the input videos as the spatial prior during training.

\noindent
\textbf{Inference.} 
Our inference process starts by performing a standard T2I process~\cite{rombach2022stablediffusion} using only the textual conditions, resulting in the high-quality spatial prior. 
Then, this spatial prior, along with the motion and appearance guidances, will be inputted into the 3D-UNet for temporal reasoning.
Next, let's explain how to construct the guidance features that correspond to the specified motion and appearance factors.
\textit{Firstly}, for a given motion factor $\gamma^{\text{m}}$, we set all elements in the vector $\tilde{\mathbf{r}}^\text{m}$ to $\gamma^{\text{m}}$, and construct the motion guidance $\mathbf{r}^{\text{m}}$ by Eq.~\ref{eq:rm}.
For a stable video, the recommended range for $\gamma^{\text{m}}$ is [300, 600].
\textit{Secondly}, for appearance guidance, we can manually construct the variation matrix $\tilde{\mathbf{r}}^\text{a}$ based on the given appearance factor $\gamma^{\text{a}}$:
%
\begin{equation}
    \scriptsize{\tilde{\mathbf{r}}^\text{a}=
    \left\{\begin{array}{cccc}
    0k + 1, & 1k + 1, & \cdots & (F-1)k+1,\\
    1k + 1, & 0k + 1, & \cdots & (F-2)k+1,\\
    \vdots & \vdots & \ddots & \vdots\\
    (F-2)k + 1, & (F-3)k + 1, & \cdots & 1k+1,\\
    (F-1)k + 1, & (F-2)k + 1, & \cdots & 0k+1,\\
    \end{array}\right\},
}
\end{equation}
where $k=\frac{-\gamma^{\text{a}}}{F-1}$. 
The variation matrix $\tilde{\mathbf{r}}^\text{a}$ is obtained by linear interpolation, resulting in smooth appearance changes between consecutive frames. 

\section{Experiments}

\subsection{Implementation Details}

\textbf{Optimization.} In this work, all modules are trained using the AdamW~\cite{loshchilov2017AdamW} optimizer with a learning rate of 5e-5. 
The weight decay is set to 0, and our default training iteration is 25,000. 
The spatial resolution of the videos is 448$\times$256. 
During the image-video joint training, the batch size for images is 512 (distributed across 2 GPUs), the number of video frames $F$ is 32, and the batch size for videos is 72 (distributed across 6 GPUs). 
Therefore, we use 8$\times$A100 GPUs to fine-tune the denoiser. 
Besides, for the pre-trained parameters from Stable Diffusion (\ie, the spatial layers), we apply a decay of 0.2 to their gradients.

\noindent
\textbf{Datasets.} The dataset used in our study consists of two types: video-text pairs and image-text pairs. 
For the video dataset, apart from the publicly available WebVid10M~\cite{bain2021webvid10m}, 
we also select a subset of aligned textual and video from our internal data, amounting to a total of 20 million video-text pairs.
The image dataset primarily consists of LAION-400M~\cite{schuhmann2021laion} and similar private image-text pairs, comprising around 60 million images.
In the ablation experiments, for efficiency, we gathered 69 commonly used imaginative prompts from recent works for testing, which will be included in our Appendix. 
For the comparison of Fr{\'e}chet Inception Distance (FID)~\cite{parmar2022fid}, Fr{\'e}chet Video Distance (FVD)~\cite{unterthiner2018fvd} and CLIP Similarity (CLIPSIM)~\cite{wu2021clipsim} metrics with state-of-the-art in Tab.~\ref{tab:msrvtt}, we evaluated the same MSR-VTT dataset~\cite{xu2016msrvtt} as previous works.

\begin{table}[t]
    \centering
    \tablestyle{4pt}{0.95}
    \begin{tabular}{c|cc|cc}
      & SL & CL  & \makecell{Temporal \\ Consistency} $\uparrow$  & CLIPSIM$\uparrow$ \\
       \shline
      ModelScope~\cite{wang2023modelscope} & \xmark &  \xmark  & 0.931 & 0.292 \\ 
      $\downarrow$ & \cmark &  \xmark  & 0.889 & 0.313\\ 
      \method & \cmark &  \cmark  & 0.944 & 0.318 \\
    \end{tabular}
    \vspace{-3mm}
    \caption{Ablation studies for Structure-Level (SL) decoupling and Content-Level (CL) decoupling.}
    \vspace{-6mm}
    \label{tab:decoupling}
\end{table}

\begin{figure*}[t]
    \centering
    \includegraphics[width=0.95\linewidth]{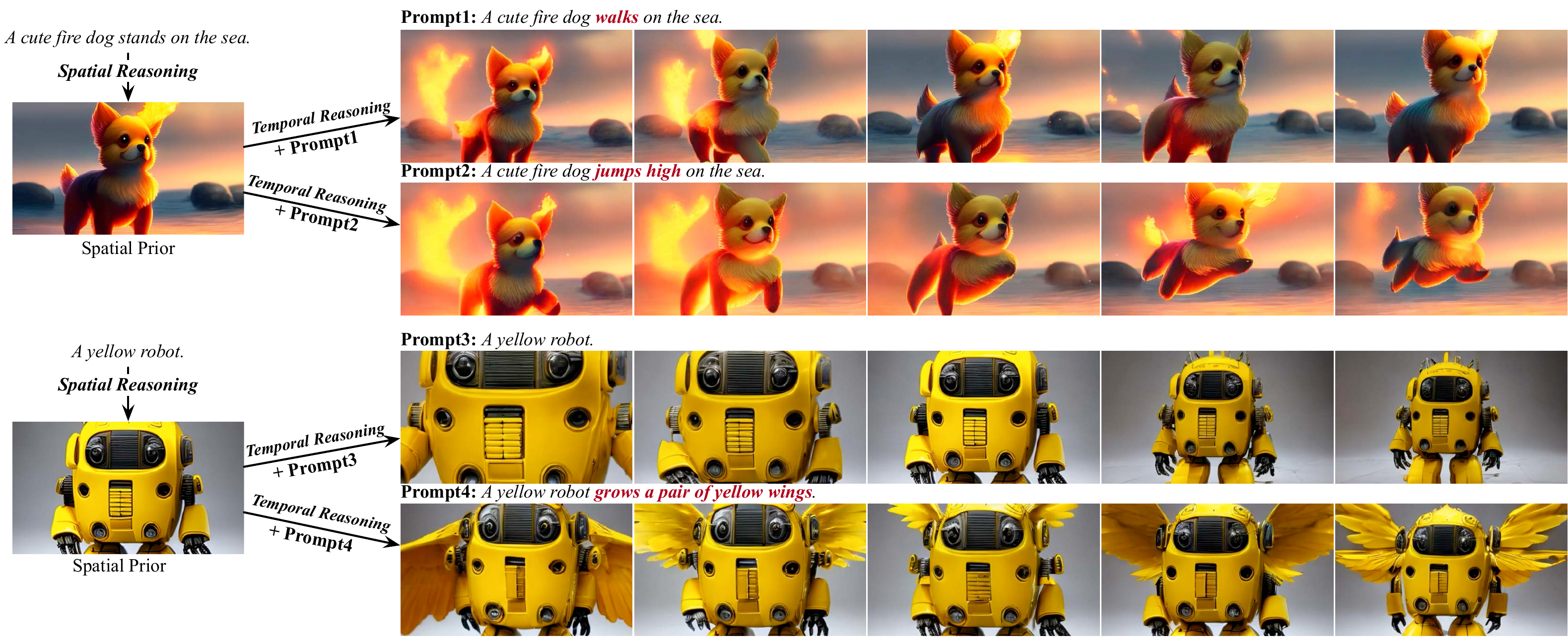}
    \vspace{-3mm}
    \caption{Combining the same spatial prior with different textual prompts allows dynamic control over the generated videos during the temporal reasoning stage.}
    \vspace{-5mm}
    \label{fig:1img_2text}
\end{figure*}

\begin{figure}[t]
    \centering
    \includegraphics[width=0.99\linewidth]{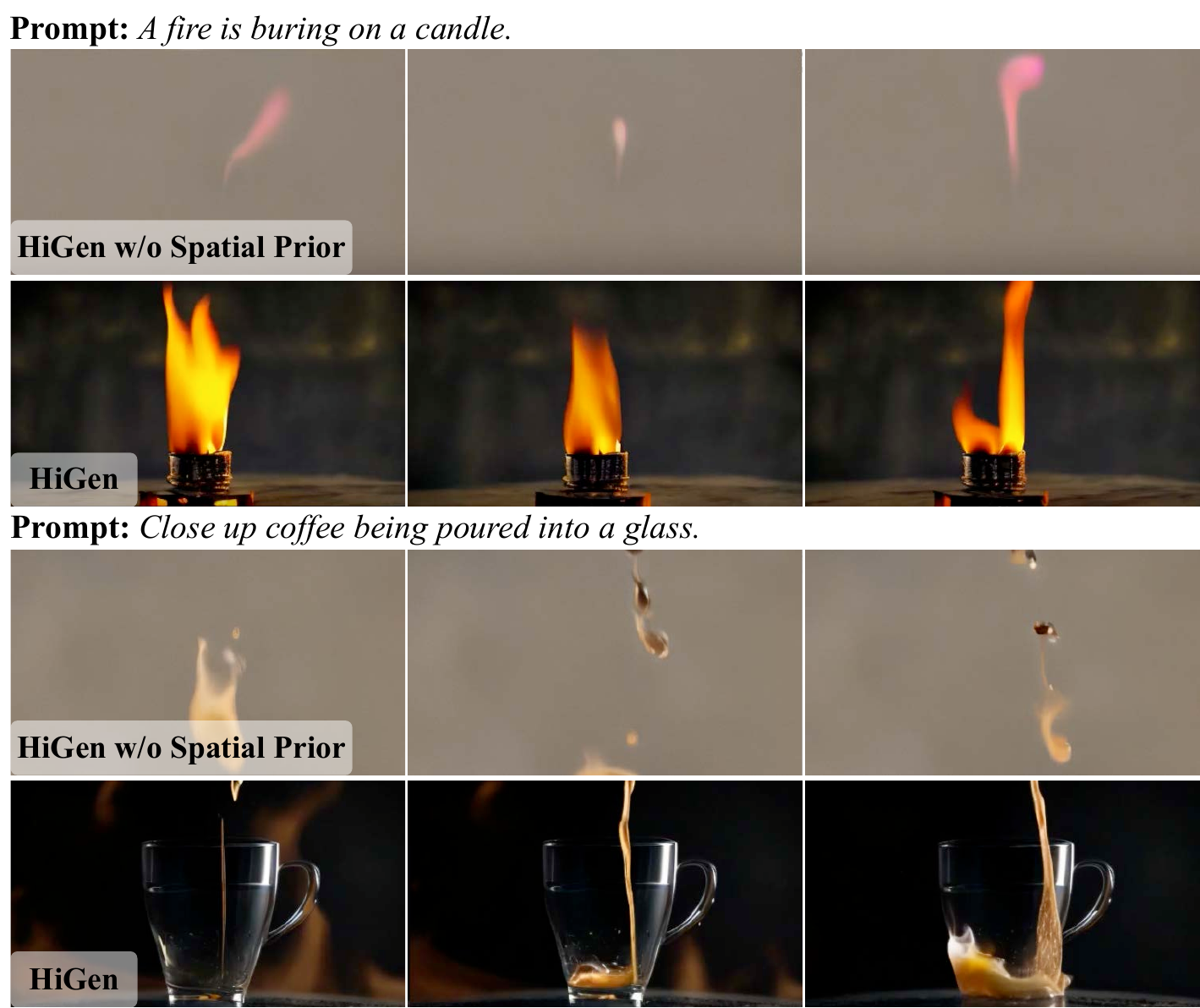}
    \vspace{-3mm}
    \caption{{Visualization for structure-level decoupling.} 
    ``HiGen w/o Spatial Prior'' refers to our temporal reasoning without inputting any spatial priors.
    }
    \vspace{-6mm}
    \label{fig:st_decoupling}
\end{figure}

\subsection{Ablation Studies} 

\noindent
In this section, we will analyze our hierarchical spatio-temporal decoupling mechanism.
Our baseline is ModelScopeT2V~\cite{wang2023modelscope}. 
Unless otherwise specified, we default to setting the motion factor $\gamma^\text{m}$ to 500 and the appearance factor $\gamma^\text{a}$ to 1.0.

\noindent
\textbf{The effect of hierarchical decoupling.} 
Comparing the first two rows of Tab.~\ref{tab:decoupling}, the structure-level decoupling significantly improves the spatial quality (\ie, CLIPSIM), but it severely compromises temporal consistency. 
The first two rows of Fig.~\ref{fig:abla1} also provide a more intuitive demonstration of this effect. 
Content-level decoupling, as shown in the third row of Tab.~\ref{tab:decoupling} and Fig.~\ref{fig:abla1}, ensures superior spatial quality and improved temporal stability of the video frames.

\noindent
\textbf{Temporal reasoning analysis.} 
In Fig.~\ref{fig:st_decoupling}, we visualize videos generated without spatial priors, showing a decoupling between temporal and spatial synthesis. 
The absence of additional spatial priors results in videos that primarily exhibit motion correlated with the text. 
Combining temporal reasoning with spatial priors reduces the complexity of video synthesis and enables high-quality results. 
Additionally, in Fig.~\ref{fig:1img_2text}, we synthesize videos using the same spatial prior but different textual prompts, observing that the temporal reasoning stage effectively utilizes the motion knowledge provided by the text prompts.

\begin{figure}
    \centering
    \includegraphics[width=0.99\linewidth]{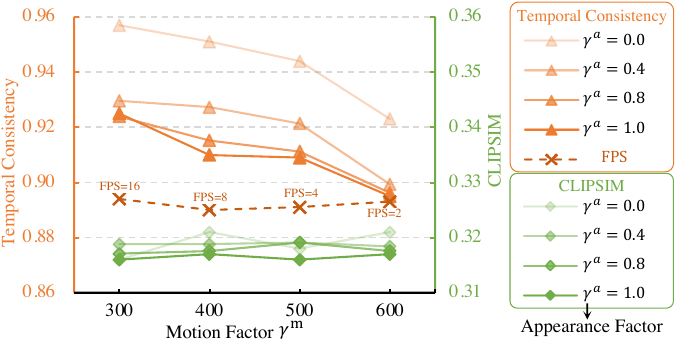}
    \vspace{-3mm}
    \caption{Parameter sensitivity analysis of the motion factor $\gamma^{\text{m}}$ and appearance factor $\gamma^{\text{a}}$.}
    \vspace{-5mm}
    \label{fig:motion_semantic_curve}
\end{figure}

\noindent
\textbf{Content-level decoupling analysis.} 
In Fig.~\ref{fig:motion_semantic_curve}, the curves demonstrate the impact of motion and appearance factors on the generated videos. 
Higher values of the motion factor (300 to 600) and appearance factor (0 to 1.0) decrease temporal consistency, while the spatial semantic remains stable according to the CLIPSIM metric. 
The dashed line represents using FPS as an alternative to our content-level decoupling strategy. Notably, changing the FPS has minimal influence on the temporal dynamics of the videos, validating the superiority of our decoupling strategy as a more effective design choice.

In addition, Fig.~\ref{fig:sec_fig} visually illustrates the impacts of these two factors. 
The motion factor governs scene movement, while the appearance factor enables diverse semantic variations in the generated videos. 
Interestingly, lower temporal consistency scores lead to livelier and more dynamic videos. 
This suggests that \textit{overly prioritizing temporal consistency may hinder the potential for vibrant and engaging videos}.

\begin{figure*}[ht]
    \centering
    \includegraphics[width=0.95\linewidth]{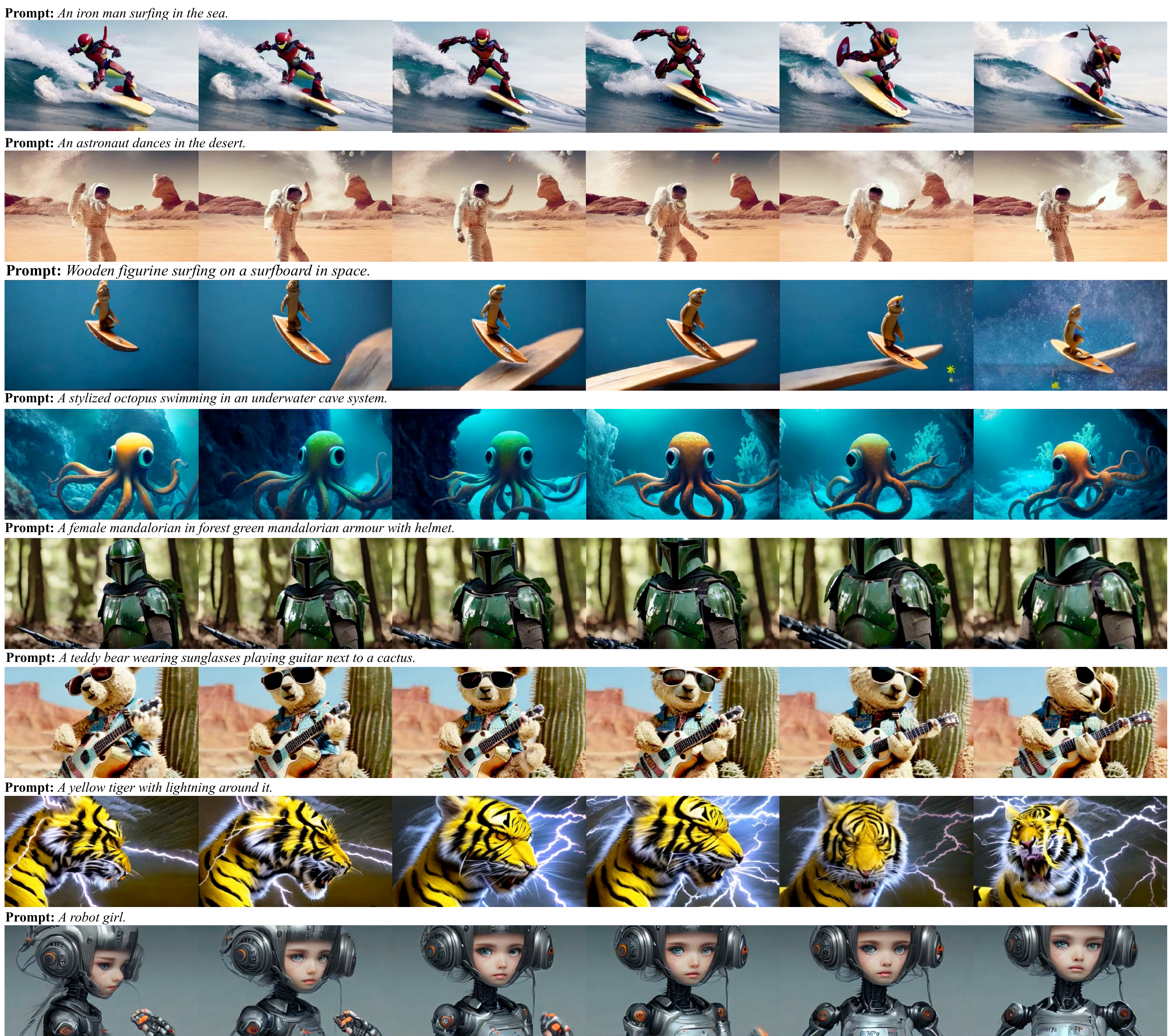}
    \vspace{-3mm}
    \caption{Sample visualization of generated videos.}
    \vspace{-4mm}
    \label{fig:case_vis}
\end{figure*}

\begin{figure}[t]
    \centering
    \includegraphics[width=0.95\linewidth]{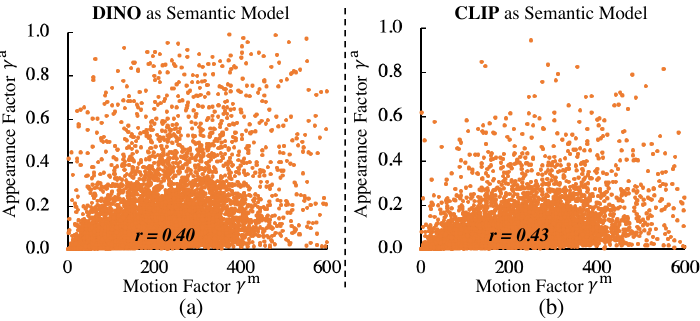}
    \vspace{-4mm}
    \caption{Correlation analysis between the motion factor and appearance factor with    DINO~\cite{caron2021dino,oquab2023dinov2} and CLIP~\cite{radford2021CLIP} as semantic models.
    Here, we measure these factors based on the first and last frames of 8000 random videos.
    }
    \vspace{-6mm}
    \label{fig:correlation}
\end{figure}

\begin{figure}[ht]
    \centering
    \includegraphics[width=0.95\linewidth]{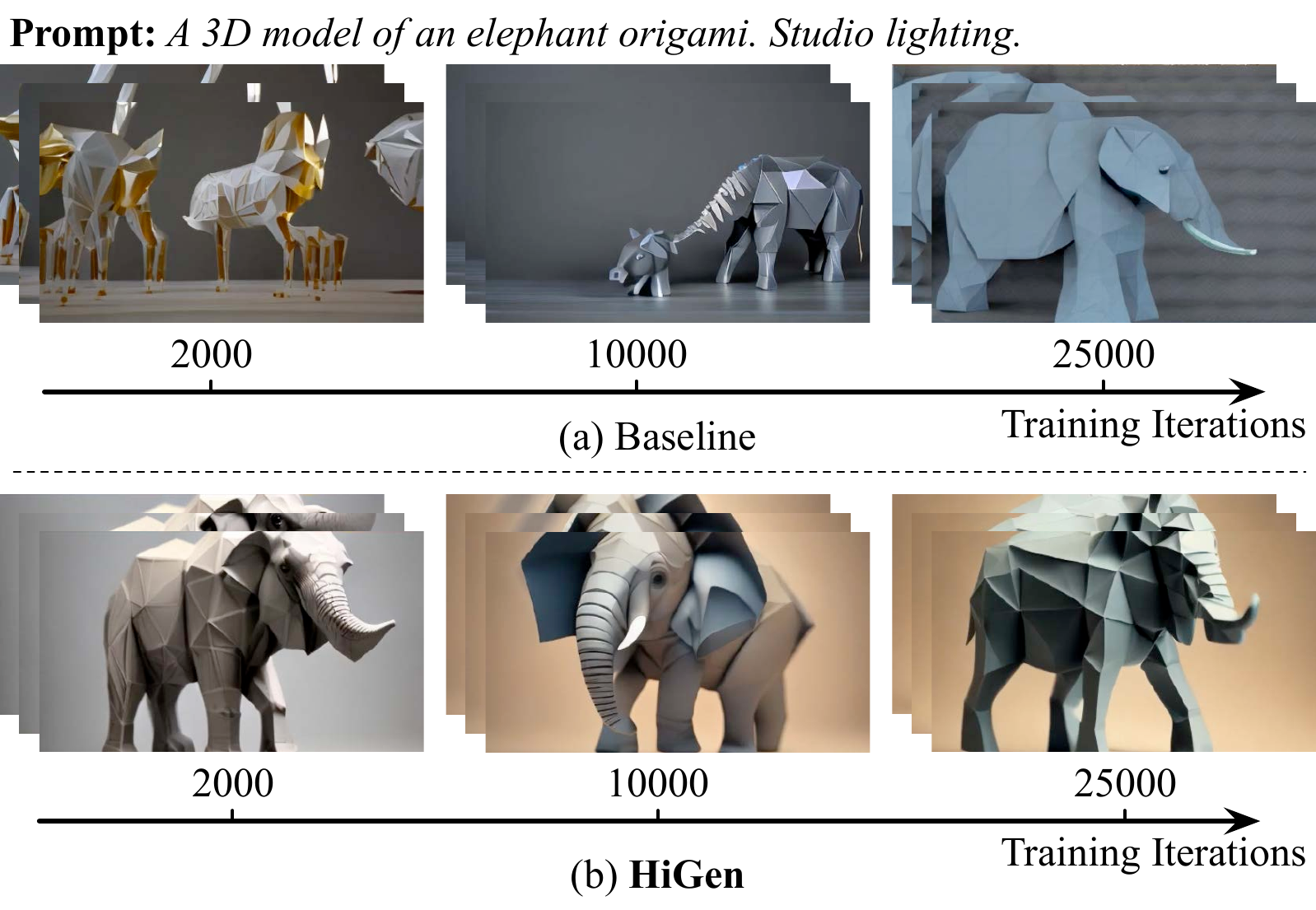}
    \vspace{-4mm}
    \caption{Comparison with baseline at various training stages.}
    \vspace{-5mm}
    \label{fig:training_iter}
\end{figure}

\begin{figure*}[ht]
    \centering
    \includegraphics[width=0.99\linewidth]{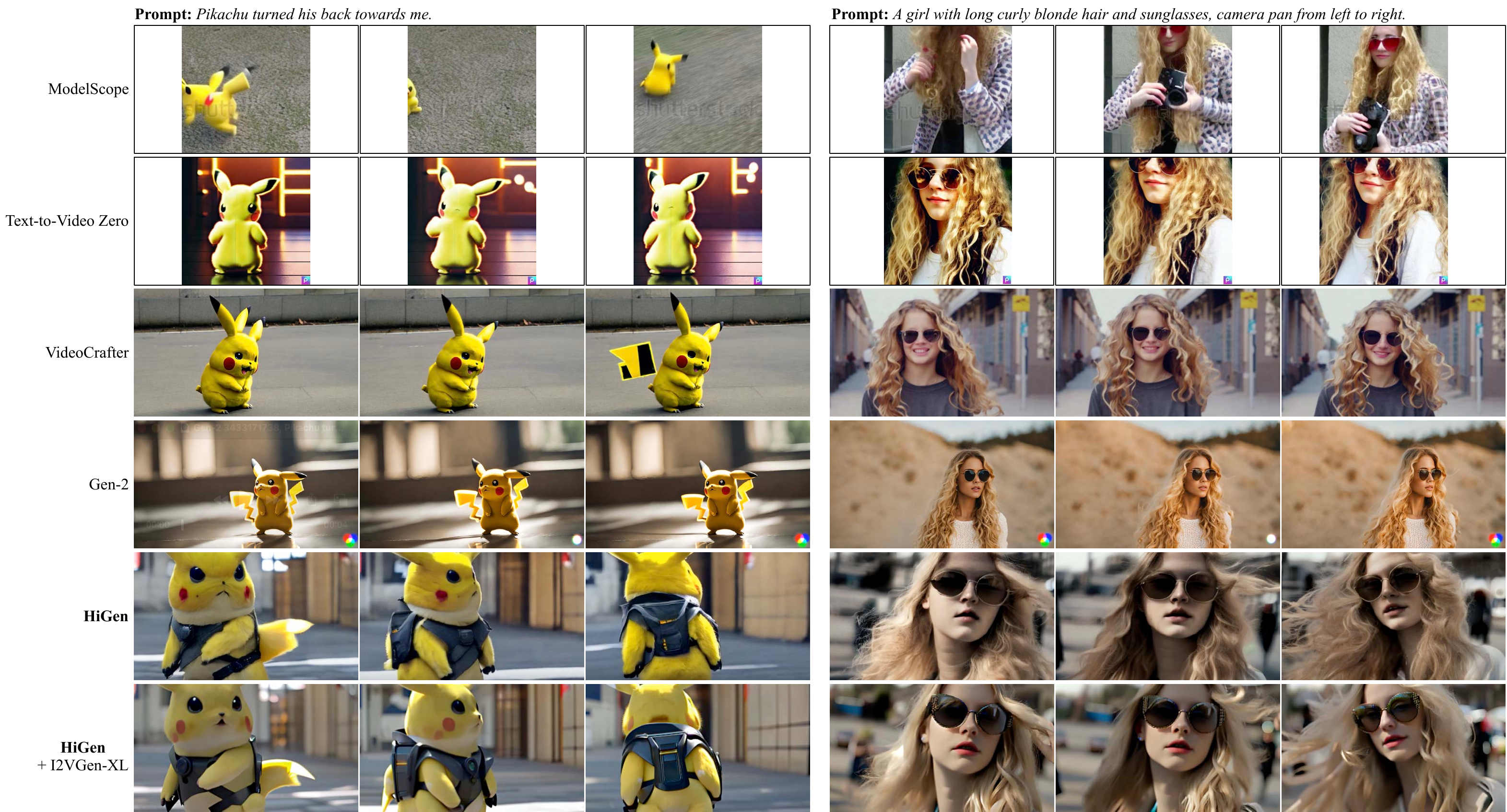}
    \vspace{-1mm}
    \caption{Qualitative comparison with ModelScopeT2V~\cite{wang2023modelscope}, Text-2-Video Zero~\cite{khachatryan2023text2video_zero}, VideoCrafter~\cite{chen2023videocrafter1} and Gen-2~\cite{esser2023gen1}. 
    In the last row, we utilize the Video-to-Video model from the open-sourced I2VGen-XL~\cite{zhang2023I2VGen} to enhance the spatial resolution of our videos, resulting in further improvement in spatial quality.}
    \vspace{-4mm}
    \label{fig:sota_vis}
\end{figure*}

\noindent
\textbf{Semantic model analysis.} 
To achieve content-level decoupling, we aim to ensure high independence between the motion and appearance factors. 
To accomplish this, we explore self-supervised models such as DINO~\cite{caron2021dino,oquab2023dinov2} and the multimodal model CLIP~\cite{radford2021CLIP} as semantic models. 
We evaluate the Pearson Correlation Coefficients (PCC) between these two factors. 
In Fig.~\ref{fig:correlation}, we observe that although the PCC between the DINO-based appearance factor and motion factor is only slightly lower (0.03) than that of CLIP, the distribution of DINO is more uniform. 
This indicates that self-supervised models, specifically DINO, are more sensitive to appearance variations. 
Based on this finding, we default to using DINO as our semantic model.

\begin{table}[t]
    \centering
    \tablestyle{4pt}{0.95}
    \begin{tabular}{c|ccc}
        Method & \makecell{Visual \\ Quality}  & \makecell{Temporal \\ Quality}  & \makecell{Text \\ Alignment} \\
        \shline
        ModelScopeT2V~\cite{wang2023modelscope} & 32.4\%& 43.2\%& 54.8\%\\ 
        Text2Video-Zero~\cite{khachatryan2023text2video_zero} & 63.6\% & 26.0\% & 53.8\%\\
        VideoCrafter~\cite{chen2023videocrafter1}  & 81.2\%& 55.2\% & 76.8\% \\ 
        \hline
        \method & \textbf{84.4}\% & \textbf{74.0}\% &\textbf{81.2}\%
    \end{tabular}
    \vspace{-3mm}
    \caption{Human evaluations with open-sourced methods.}
    \vspace{-6mm}
    \label{tab:human_eval}
\end{table}
\noindent
\textbf{Training efficiency.} The structure-level decoupling of spatial and temporal aspects mitigates the difficulties in joint spatio-temporal denoising. 
Fig.~\ref{fig:training_iter} compares the generated videos at different iterations with the baseline method. 
It is clear that \method consistently outperforms the baseline regarding visual quality throughout various training stages.

\noindent
\textbf{More visualizations.}
Fig.~\ref{fig:case_vis} demonstrates the generation of 8 different styles of videos, such as humans, animals, and marine settings. 
The generated videos using \method showcase consistent, high-quality frames comparable to Stable Diffusion-generated images. 
When played in sequence, these frames exhibit both smooth temporal content and diverse semantic variations, enhancing the richness and vividness of the videos.

\noindent
\textbf{Human evaluations.} In Tab.~\ref{tab:human_eval}, we conducted a human evaluation of three recent open-source methods, considering spatial, temporal, and textual aspects. 
Notably, \method exhibits the most substantial improvement in temporal performance, surpassing VideoCrafter~\cite{chen2023videocrafter1} by 18.8\% (increasing from 55.2\% to 74.0\%). 
These findings further reinforce the superiority of our approach.

\subsection{Comparison with State-of-the-art}

Tab.~\ref{tab:msrvtt} compares \method with existing approaches using FID, FVD, and CLIPSIM metrics on MSR-VTT~\cite{xu2016msrvtt}. 
Our method shows significant improvements in FID and FVD metrics. 
However, as noted in previous works~\cite{podell2023sdxl}, these metrics may not accurately represent the generated quality. 
To further evaluate, we visually compare our results with recent state-of-the-art methods in Fig.~\ref{fig:sota_vis}. 
It is evident that our \method achieves a better balance between spatial quality and temporal motion in the generated videos.

\begin{table}[t]
    \centering
    \tablestyle{4pt}{0.95}
    \begin{tabular}{c|ccc}
      Method & FID $\downarrow$& FVD$\downarrow$  & CLIPSIM$\uparrow$ \\
      \shline
      CogVideo (English)~\cite{cogvideo} & 23.59 & 1294 & 0.2631 \\
      Latent-Shift~\cite{an2023latent_shift} & 15.23  & - & 0.2773 \\
       Make-A-Video~\cite{singer2022make_a_video}  & 13.17 &-& \textbf{0.3049} \\
        Video LDM~\cite{blattmann2023videoLDM}  & - &-& 0.2929 \\
        MagicVideo~\cite{zhou2022magicvideo}  & - &998& - \\
        VideoComposer~\cite{wang2023videocomposer} & 10.77 &580&0.2932 \\
        ModelScopeT2V~\cite{wang2023modelscope} & 11.09 & 550 &  0.2930\\
        
        PYoCo~\cite{ge2023pyoco} & 9.73 & - & - \\
        \hline
        \method & \textbf{8.60} & \textbf{406} & {0.2947} \\
    \end{tabular}
    \vspace{-3mm}
    \caption{T2V generation performance on MSR-VTT~\cite{xu2016msrvtt}.}
    \vspace{-6mm}
    \label{tab:msrvtt}
\end{table}

\section{Discussions}

This work presents \method, a diffusion model-based approach for video generation that decouples spatial and temporal factors at both the structure-level and content-level. 
With a unified denoiser, \method generates spatially photorealistic priors and temporally coherent motions, while extracting subtle cues from the video content to express appearance and motion changes for denoising guidance. 
Through this design, \method successfully reduces the complexity of T2V task, synthesizes realistic videos with semantic accuracy and motion stability, and outperforms state-of-the-art T2V methods in extensive experiments.

\noindent
\textbf{Limitations.} Due to limitations in computational resources and data quality, our \method's ability to generate object details lags behind that of current image synthesis models. Additionally, accurately modeling human and animal actions that adhere to common sense proves challenging, particularly in cases of substantial motion. 
To address these challenges, our future research will delve into improving model design and data selection.

\noindent
\textbf{Acknowledgement.}
This work is supported by the National Natural Science Foundation of China under grant U22B2053 and 62176097, and by Alibaba DAMO Academy through Alibaba Research Intern Program.
{
    \small
    \bibliographystyle{ieeenat_fullname}
    \bibliography{main}
}

\newpage
\appendix
\renewcommand{\thetable}{A\arabic{table}}
\renewcommand{\thefigure}{A\arabic{figure}}
\renewcommand{\theequation}{A\arabic{equation}}

\section*{Overview}

%
In this supplementary material, we provide experiments related to the semantic model and features for motion and appearance guidance in Sec.~\ref{sec:exp}. 
Lastly, in Tab.~\ref{tab:prompts1} and Tab.~\ref{tab:prompts2}, we provide the 69 text prompts used in our ablation experiments.

\begin{figure}
    \centering
    \includegraphics[width=0.99\linewidth]{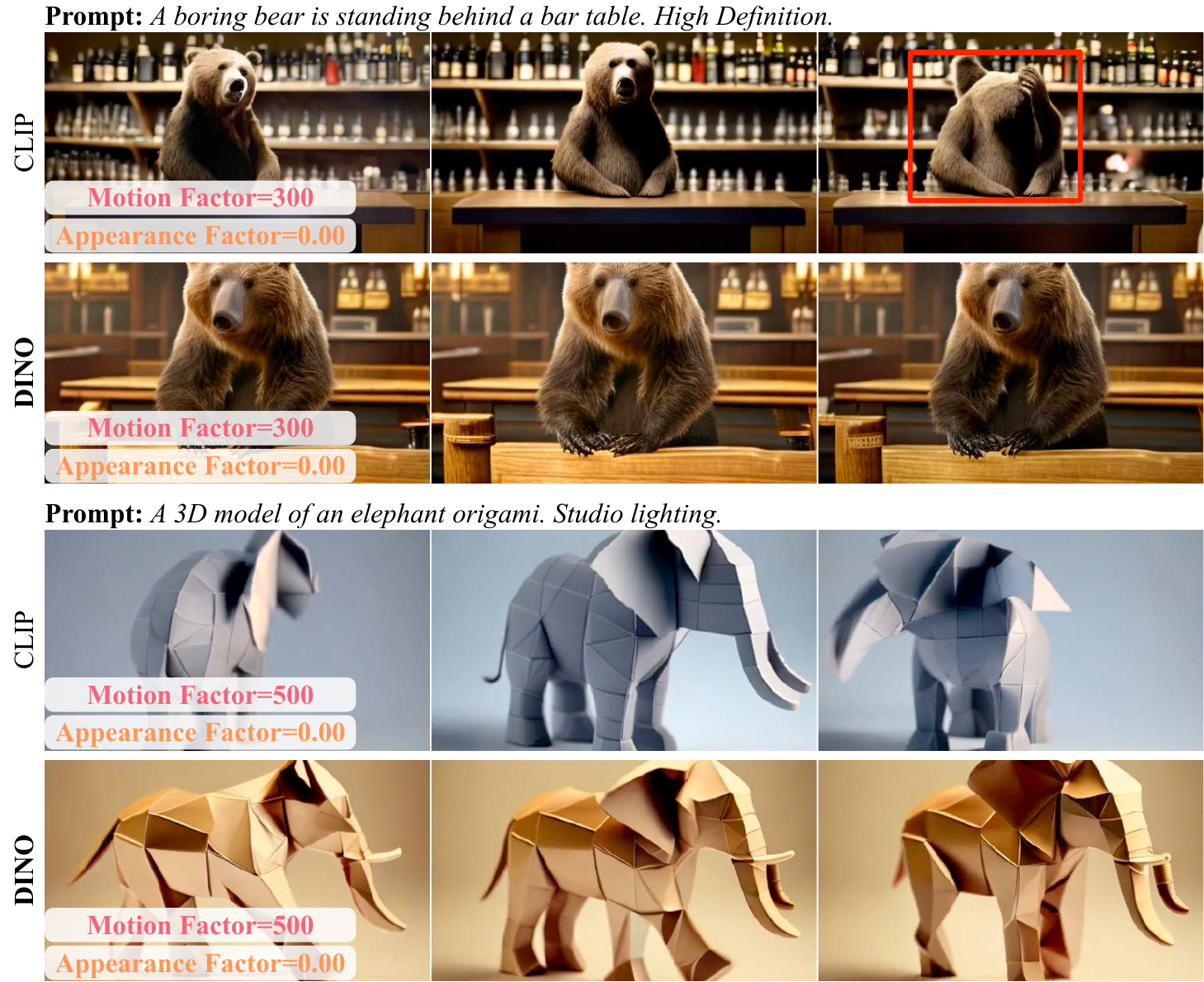}
    \caption{Visualization for different semantic models. In these four generated videos, we set the appearance factor $\gamma^{\text{a}}$ to zero, indicating that the entire video should have minimal visual changes.}
    \label{fig:clip_dino_vis}
\end{figure}

\section{More Experiments}
\label{sec:exp}
\textbf{Different semantic models.} In Fig.~\ref{fig:clip_dino_vis}, we visualize the generated results when using DINO~\cite{oquab2023dinov2} and CLIP~\cite{radford2021CLIP} as semantic models in the appearance analysis. 
It can be observed that videos generated using DINO as the semantic model exhibit better temporal stability, whereas CLIP may result in unexpected and irrational abrupt changes.
Fig.{\color{red}10} in the main paper also supports this observation by demonstrating a lower correlation between appearance and motion factors obtained from DINO features, thereby enabling better independent control.

\noindent
\textbf{How to generate motion and appearance guidance?}
In Fig.{\color{red}3} of the main paper, we default to using a vector composed of $F-1$ frame difference elements to generate motion guidance, while a similarity matrix is used to generate appearance guidance. 
The reason behind this choice is that frame difference cannot capture the motion information between frames with significant visual differences, whereas a semantic model can effectively model the appearance correlation between any pair of video frames. 
In Tab.~\ref{tab:different_control}, we quantitatively analyze the combinations of motion and appearance guidance using vector-based and matrix-based methods. 
We conducted evaluations with three different motion factors and semantic factors for each combination, and then measured changes in terms of Temporal Consistency and CLIPSIM. 
It can be observed that different combinations exhibit similar spatial quality for the generated videos (\textit{i.e.}, minimal changes in CLIPSIM), but using frame difference vectors for motion guidance and similarity matrices for appearance guidance leads to more significant temporal variations (\ie, $\pm$ 0.0276).

\begin{figure*}
    \centering
    \includegraphics[width=0.95\linewidth]{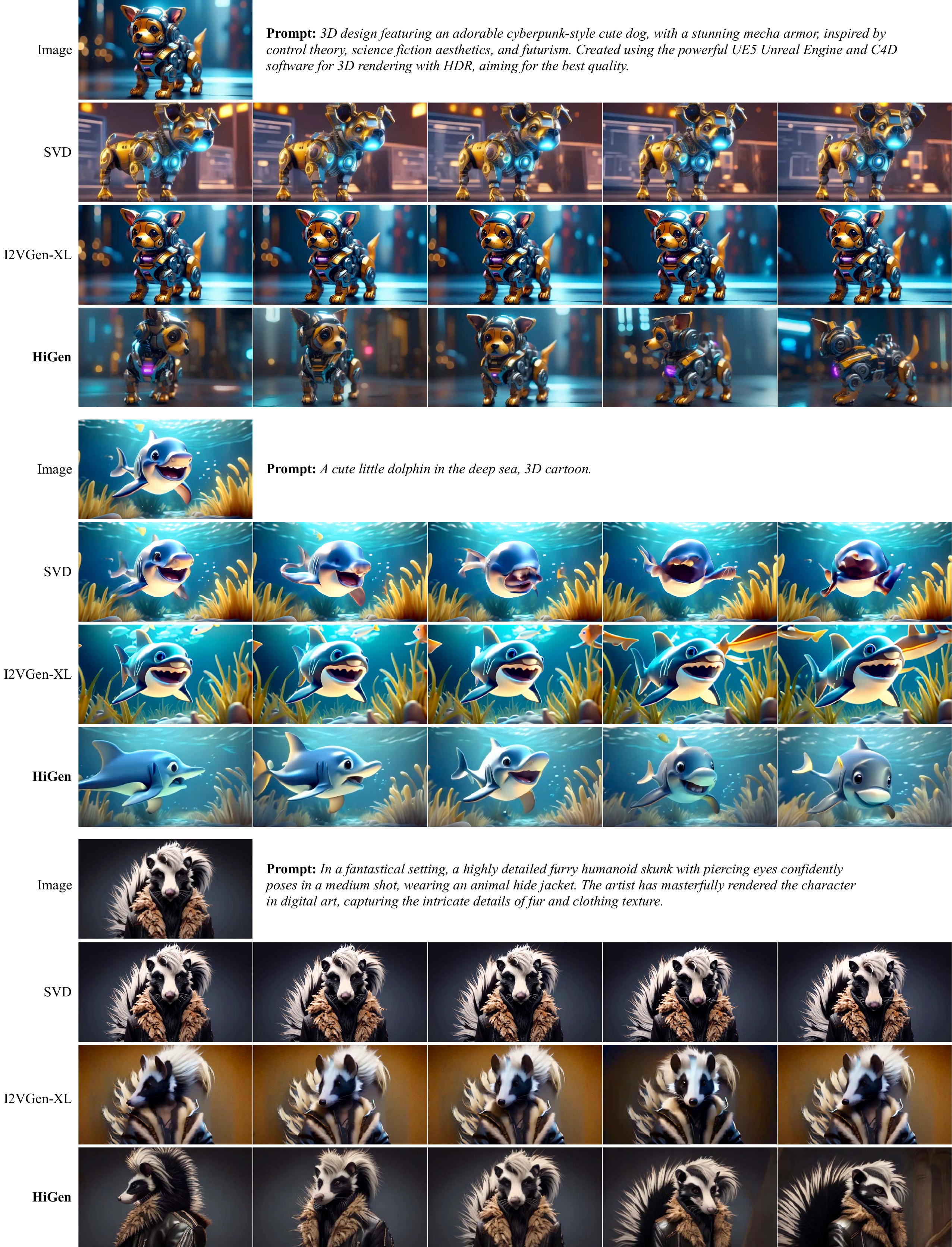}
    \caption{Comparison with advanced image-to-video methods, \ie, Stable Video Diffusion~\cite{svd} and I2VGen-XL~\cite{zhang2023I2VGen}.}
    \label{fig:appendix_i2v}
\end{figure*}

\noindent
\textbf{Comparison with image-to-video approaches.}
In Fig.~\ref{fig:appendix_i2v}, we compare \method with state-of-the-art image-to-video generation methods. 
The images are generated using advanced text-to-image methods such as Midjourney.
We directly incorporate these images as spatial priors in the temporal reasoning step. 
It can be observed that, compared to Stable Video Diffusion~\cite{svd} and I2VGen-XL~\cite{zhang2023I2VGen}, the videos generated by \method exhibit significantly more vivid temporal dynamics and creativity.

\begin{table}[t]
    \centering
    \tablestyle{4pt}{0.95}
    \begin{tabular}{cc|cc}
        \makecell{Motion \\ Guidance}& \makecell{Appearance \\ Guidance} & \makecell{Temporal \\ Consistency} &CLIPSIM \\
         \shline
         Vector & Vector & 0.9314 $\pm$ 0.0154 & 0.3171 $\pm$ 0.0015\\
         Matrix & Matrix & 0.9380 $\pm$ 0.0198 & 0.3165 $\pm$ 0.0010\\
         Matrix & Vector & 0.9392 $\pm$ 0.0167 & 0.3159 $\pm$ 0.0017\\
         Vector & Matrix & 0.9367 $\pm$ \textbf{0.0276} &0.3166 $\pm$ 0.0013\\
    \end{tabular}
    \caption{Different methods for generating motion and appearance guidance.}
    \label{tab:different_control}
\end{table}

\begin{table*}[ht]
    \centering
    \small
    \begin{tabular}{l}
    \shline
        1.  \textit{Close up of grapes on a rotating table. High Definition}\\
        2. \textit{Raw fresh beef meat fillet on a wooden plate with dill}\\
        3. \textit{Close up coffee being poured into a glass. Slow motion}\\
        4. \textit{Close-up milky liquid being poured. slow motion}\\
        5. \textit{A waving flag close up realistic animation seamless loop}\\
        6. \textit{Face of happy macho mature man looking at camera}\\
        7. \textit{Face of happy macho mature man smiling}\\
        8. \textit{A girl is looking at the camera smiling. High Definition}\\
        9. \textit{Woman in sunset}\\
        10. \textit{Young girl eye macro, shot in raw, 4k}\\
        11. \textit{Blue sky clouds timelapse 4k time lapse big white clouds cumulus growing cloud formation sunny weather background}\\
        12. \textit{Campfire at night in a snowy forest with starry sky in the background}\\
        13. \textit{Ocean waves hitting headland rocks pacifica california slow motion}\\
        14. \textit{There is a table by a window with sunlight streaming through illuminating a pile of books}\\
        15. \textit{Beautiful sexy lady in elegant white robe. close up fashion portrait of model indoors. beauty blonde woman}\\
        16. \textit{Fire burning in a forest}\\
        17. \textit{Wildfire in mountain of thailand (pan shot)}\\
        18. \textit{Fireworks}\\
        19. \textit{Melting pistachio ice cream dripping down the cone}\\
        20. \textit{A 3D model of an elephant origami. Studio lighting}\\
        21. \textit{Strawberry close-up on a black background swinging, slow motion. water flows down the berry}\\
        22. \textit{A spaceship is landing.}\\
        23. \textit{A giant spaceship is landing on mars in the sunset. High Definition}\\
        24. \textit{Drone flythrough of a tropical jungle covered in snow. High Definition}\\
        25. \textit{Fog at the end of the path in the summer-autumn forest. nobody present. scary scene. peaceful. quiet}\\
        26. \textit{Cars running on the highway at night}\\
        27. \textit{A man is riding a horse in sunset}\\
        28. \textit{close up of a clown fish swimming. 4K}\\
        29. \textit{a boring bear is standing behind a bar table. High Definition}\\
        30. \textit{Beautiful pink rose background. blooming rose flower rotation, close-up}\\
        31. \textit{A cat eating food out of a bowl, in style of Van Gogh}\\
        32. \textit{Wide shot of woman worker using welding machine on her work in site construction.}\\
        33. \textit{celebration of christmas}\\
        34. \textit{A fire is burning on a candle.}\\
        35. \textit{Pov driving high rural mountain country road snow winter blue sky nature environment sierra nevada usa}\\
        36. \textit{Lid taken off gift box with puppy inside on table top with holiday gifts.}\\
        37. \textit{Silhouette of retired caucasian american couple enjoying the sunrise having kayaking trip on the lake outdoors red dragon}\\
        38. \textit{Aerial autumn forest with a river in the mountains.}\\
        39. \textit{London, uk - november 16,2014:traffic. buses and cars in baker street move slowly in london england. congested city traffic}\\
        40. \textit{A corgi is swimming fastly}\\
        41. \textit{There is a table by a window with sunlight streaming through illuminating a pile of books}\\
        42. \textit{A glass bead falling into water with a huge splash. Sunset in the background}\\
        43. \textit{Aerial autumn forest with a river in the mountains}\\
        44. \textit{astronaut riding a horse}\\
        45. \textit{A clear wine glass with turquoise-colored waves inside it}\\
        46. \textit{A bear dancing and jumping to upbeat music, moving his whole body}\\
        47. \textit{A bigfoot walking in the snowstorm}\\
        48. \textit{An iron man surfing in the sea}\\
        49. \textit{Filling a glass with warm coffee}\\
        50. \textit{3d fluffy Lion grinned, closeup cute and adorable, long fuzzy fur, Pixar render}\\

    \shline
    \end{tabular}
    \caption{Prompts Part-I.}
    \label{tab:prompts1}
\end{table*}

\begin{table*}[t]
    \centering
    \small
    \begin{tabular}{l}
    \shline
        51. \textit{A big palace is flying away, anime style, best quality}\\
        52. \textit{A teddy bear is drinking a big wine}\\
        53. \textit{A giant spaceship is landing on mars in the sunset. High Definition}\\
        54. \textit{A happy elephant wearing a big birthday hat walking under the sea, 4k}\\
        55. \textit{Albert Einstein washing dishes}\\
        56. \textit{Blue sky clouds timelapse 4k time lapse big white clouds cumulus growing cloud formation sunny weather background}\\
    57. \textit{drone flythrough interior of sagrada familia cathedral}\\
        58. \textit{Close up of grapes on a rotating table. High Definition}\\
        59. \textit{A stunning aerial drone footage time lapse of El Capitan in YosemiteNational Park at sunset}\\
        
        60. \textit{Aerial autumn forest with a river in the mountains}\\
        61. \textit{An astronaut dances in the desert}\\
        62. \textit{Blue sky clouds timelapse 4k time lapse big white clouds cumulus growing cloud formation sunny weather background}\\
        63. \textit{Beautiful pink rose background. blooming rose flower rotation, close-up}\\
        64. \textit{Fog at the end of the path in the summer-autumn forest. nobody present. scary scene. peaceful. quiet}\\
        65. \textit{A beautiful sunrise on mars, Curiosity rover. High definition,timelapse, dramatic colors}\\
        66. \textit{Van Gogh is smoking}\\
        67. \textit{A shiny golden waterfall flowing through glacier at night}\\
        68. \textit{A teddy bear painting a portrait}\\
        69. \textit{Fog at the end of the path in the summer-autumn forest. nobody present. scary scene. peaceful. quiet}\\
    \shline
    \end{tabular}
    \caption{Prompts Part-II.}
    \label{tab:prompts2}
\end{table*}
        
\section{Prompts}
\label{sec:prompts}
Finally, we will provide the 69 text prompts that were used in Tables {\color{red}1}, {\color{red}2}, and Figure {\color{red}8} of our main paper.


\end{document}